\documentclass[10pt,twoside,twocolumn,english]{journal}
\usepackage{mathptmx}

\usepackage[T1]{fontenc}
\usepackage[latin9]{inputenc}
\usepackage[letterpaper]{geometry}
\usepackage{color}
\usepackage{babel}
\usepackage{float}
\usepackage{graphicx}
\usepackage[unicode=true,
 bookmarks=true,bookmarksnumbered=false,bookmarksopen=false,
 breaklinks=false,pdfborder={0 0 1},backref=false,colorlinks=false]
 {hyperref}
\hypersetup{pdftitle={ROSgeoregistration: Aerial Multi-spectral Image Simulator for the
Robot Operating System},
 pdfauthor={Andrew R. Willis, Kevin M. Brink and Kathleen Dipple},
 allcolors=blue}

\makeatletter

\providecommand{\tabularnewline}{\\}

\renewcommand{\textendash}{--}

\@ifundefined{showcaptionsetup}{}{%
 \PassOptionsToPackage{caption=false}{subfig}}
\usepackage{subfig}
\makeatother

\begin{document}


%

\author{
    Andrew R. Willis,
    Kevin Brink,
    and~Kathleen Dipple

\thanks{A. Willis is with the Department of Electrical and Computer Engineering, Univ. of North Carolina at Charlotte, Charlotte,
NC, 28223 USA e-mail: arwillis@uncc.edu}
\thanks{K. Brink and K. Dipple are with the Air Force Research Laboratory, Eglin AFB, FL, 32547 USA e-mail: kevin.brink@afresearchlab.mil and kathleen.dipple.1@us.af.mil}
}

\title{ROS georegistration: Aerial Multi-spectral Image Simulator for the Robot Operating System}
\maketitle

\begin{abstract}
This article describes a software package called ROSgeoregistration 
intended for use with the Robot Operating System (ROS) and the Gazebo 3D simulation environment. ROSgeoregistration provides tools for the simulation, test and deployment of aerial georegistration algorithms and is made available with a link provided in the paper. 
A model creation package is provided which downloads
multi-spectral images from the Google Earth Engine database and, if
necessary, incorporates these images into a single, possibly very
large, reference image. Additionally a Gazebo plugin which uses the real-time sensor
pose and image formation model to generate simulated imagery using
the specified reference image is provided along with related plugins for UAV relevant data. The novelty of this work is threefold:
(1) this is the first system to link the massive multi-spectral imaging
database of Google's Earth Engine to the Gazebo simulator, (2) this
is the first example of a system that can simulate geospatially and
radiometrically accurate imagery from multiple sensor views of the
same terrain region, and (3) integration with other UAS tools creates
a new holistic UAS simulation environment to support UAS system and
subsystem development where real-world testing would generally be
prohibitive. Sensed imagery and ground truth registration information
is published to client applications which can receive imagery synchronously
with telemetry from other payload sensors, e.g., IMU, GPS/GNSS, barometer,
and windspeed sensor data. To highlight functionality, we demonstrate
ROSgeoregistration for simulating Electro-Optical (EO) and Synthetic
Aperture Radar (SAR) image sensors and an example use case for developing
and evaluating image-based UAS position feedback, i.e., pose for
image-based Guidance Navigation and Control (GNC) applications.
\end{abstract}


\section{Introduction}

Visual sensors have long been used to aid navigation in many autonomous
and semi-autonomous flights systems. Concepts like visual odometry
\cite{VO_amidi1999} and simultaneous localization and mapping (SLAM)
\cite{SLAM_Bryson2008} can be used to help bound drift inherent in
inertial navigation systems \cite{INSDRIFT_Wang2014}. Other more
recent approaches such as the multi-state constrained Kalman filter
(MSCKF) \cite{MSCKF_mourikis2007} and pose-graph optimizations as
an alternative to filter-based SLAM \cite{PoseGraphSLAM_Youyang2020}
have been introduced to further improve estimation accuracy, robustness,
and/or computational efficiency. Some approaches have leveraged georegistered
reference maps and try to match imagery taken in flight to these reference
maps to provide the estimator with measurements or factors such as
bearings-to-known-features \cite{VethBook} or even direct position
updates\textbf{ }via sensor pose recovery from corresponding image
points \cite{Hartley:2003:MVG:861369}, or similar. Regardless of
the exact approach, there is a long legacy of visual aiding for autonomous
navigation that is continuing to expand with machine learning and
related developments \cite{ML4VisionNav} and an exploration beyond
standard electro-optical (EO) and infra red (IR) and into a wide range
of sensing modalities including passive millimeter wave \cite{PMMW_suss1989},
synthetic aperture radar (SAR) imaging \cite{SAR_montenbruck2008},
other RF-based techniques, and more.

This paper presents an academic research and development (R\&D) resource
which provides a flight environment modeling and simulation (M\&S)
capability for unmanned aerial systems (UAS) including fixed-wing
and multi-rotor systems. Specifically, this work places an emphasis
on providing EO and SAR image M\&S capabilities to compliment inertial
measurement unit (IMU) data and associated truth data to facilitate
academic navigation and associated image processing R\&D efforts.
The intent of this work is to help address the difficulty many researchers
face regarding GPS-denied/degraded navigation R\&D, namely access
to relevant data. 

The software packages here provide means to generate the relevant
data (images, IMU, truth) for desired flight profiles, assuming a
reasonable height above ground. The environment also provides a means
to develop and test navigation relevant vision processing concepts
while using stand-ins (e.g., UAV and camera truth data) where desired,
or to run a complete image-aided navigator pipeline in real-time within
the ROS/Gazebo environment, real-time playback, or with non-real-time
log files. 

The code-base for ROSgeoregistration can be found out: \cite{ROSgeoregistration} and the authors hope the visual simulation environment, with its tie-ins
to effective and efficient vehic\textcolor{black}{le modeling and
related simulation capabilities, provides an effective launching
point for additional development, both for image processing and navigation
within this environment, but also for expanding the underlying simulation
capability as well. As an example, the utility
of this simulator is demonstrated on SAR and EO georegistration applications and several
other potential use cases are highlighted. }

\textcolor{black}{The contributions of this work are: }
\begin{enumerate}
\item Linking the massive multi-spectral imaging database of Google's Earth
Engine to the Gazebo simulator, 
\item Simulating geospatially and radiometrically accurate imagery from
multiple sensor views for the same terrain region,
\item Integration with other quality UAS tools within a single UAS simulation
environment to broadly support UAV and image/signal processing development. 
\end{enumerate}
The code used in this paper can be found at: \url{https://github.com/uncc-visionlab/rosgeoregistration}. 

\textcolor{black}{The rest of the paper is organized as follows: Section
\ref{sec:Prior-Work} describes prior work on simulation environments
and related work which can benefit from such tools. Methodology, including
integration of core components and new developments, is discussed
in section \ref{sec:Methodology}. Section \ref{sec:Results} provides
the results of this work including examples of the data content and
imagery produced and example use case. Finally, a conclusion is provided
in section \ref{sec:Conclusion}. }

\section{\textcolor{black}{\label{sec:Prior-Work} Prior and Related Work}}

This section briefly describes relevant prior work pertaining to simulation
for use in image processing development, especially with ties into
navigation, and some of the relevant image processing and navigation
concepts this simulation environment can help facilitate. 

\subsection{Vehicle Simulation and Image Generation}

A host of prior simulation environments and tools have been made available
by the community, however, thorough literature searches 
 found no widely available simulator that integrates flight simulation
capability at this level (ROSflight \cite{Rosflight}, ROSplane \cite{Rosplane},
ROScopter \cite{ROScopter}, etc.) with real EO imagery and real SAR
measurements.

Some very useful developments provide a ROS environment with reasonable
flight dynamics, often for quadrotors \cite{TumSim}, some of which
have outdoor vision environments \cite{Hector}. These resources have
the advantage of allowing low altitude flight, but are they generally
not based on real imagery and would not allow for multi-spectral
implementations. 

Other vision simulators such as AirSim \cite{AirSim} and WorldWind
\cite{WorldWind} provide image generation capability. The latter includes
camera models leveraging real image data, but has a limited interface
and does not appear to have been integrated with ROS vehicle and sensor
modeling. The lack of ``full stack'' vehicle and camera,
especially realistic camera data, continues to lead to data set generation
\cite{Jurevicius2019,DataSetsETHZ}. These data sets are highly valuable,
especially those with non-simulated data; however the reality is most
researchers do not have control over the content, the format, the
flight plans, and the data sets are generally fairly limited (by necessity).
There is a need for a more customizable, and potentially scalable
approach. 

To the best of the authors' knowledge, this is the first ``full stack''
simulation with real EO imagery and real SAR measurements, especially
for the case were realistic models for flight vehicle dynamics, autopilot,
and additional sensing (IMU, etc.) is also readily available to support
a wide range of developments.

\subsection{Image Processing and Navigation}

A wide range of image processing capabilities are already available which can used in conjunction with the image generation capabilities provided here. This includes the widely used OpenVC \cite{OpenCV} and existing work specifically look at matching EO images to EO satellite data \cite{EO2EO}, frame to
frame feature tracking \cite{FeatureTracking}, as well as classic structure
from motion \cite{StrucFromMotion}. 

A wide range of past work has looked and image processing to aid navigation
\cite{MSCKF_mourikis2007,VethBook} and some specifically leveraged
ROS and associated simulation capabilities for development prior to
hardware implementations \cite{EllingsonNavigation2020}. The environment
presented in this paper includes highly flexible image model framework,
discussed in section \ref{sec:Methodology}, and is intended to increase
the fidelity of the referenced prior works and allow their expansion
to include seasonal variations, multi-spectral matching, e.g., SAR-to-EO \cite{Paul_2020} or SAR-to-IR \cite{EO2IR_Huang2012}, and
more. 

\subsection{SAR Simulation Considerations}

This simulator provides researchers access to investigate UAS imaging
technologies not otherwise possible due to prohibitive SWaP-C (Size,
Weight, Power and Cost) requirements. As such, ROSgeoregistration
makes it possible for researchers to explore how these data can enhance
UAS capabilities and potentially provide insight on how satellite
imagery can be used to facilitate UAS applications.

As an example, consider the Sentinel-1 C-band SAR data made available
through this effort. In this case, while there has been some work
on SAR simulation \cite{Auer_2016}, there is no open-source solutions
that enable researchers to generate realistic SAR imagery. Further,
federal regulations, platform requirements and the procurement and
maintenance costs required to deploy these sensors makes real-world
systems prohibitively difficult for most research teams. These barriers
to entry discourage SAR image research and slow the development of
this technology which shows significant promise for GNC applications
since SAR sensors can ``see through'' clouds, smoke, rain and other
atmospheric conditions that block the view of visible light sensors.
ROSgeoregistration provides researchers capabilities to leverage SAR
data from existing databases and process this data to represent degraded
SAR processing due to INS errors which has applications in SAR image
formation algorithms for static \cite{SARerror}, or moving targets
\cite{Sar_addTarget} and a wide range of other contexts.

\section{\label{sec:Methodology}Methodology}

ROSgeoregistration integrates with existing UAS simulation tools,
e.g., ROSflight, ROSplane, ROScopter, to allow a rich variety of sensor
fusion applications that may fuse IMU, GPS/GNSS, barometer, and wind-speed
sensor data with image registration results. ROSgeoregistration consists
of three components/contributions: 
\begin{enumerate}
\item A \emph{Google Earth Engine client} to construct multi-spectral terrain
models for Gazebo (\S~\ref{subsec:Google-Earth-Engine}),
\item A \emph{ROS+Gazebo plugin} that generates sensor images from the terrain
models (\S~\ref{subsec:ROS+gazbeo-plugin}),
\item A collection of \emph{core components} that integrate with the ROS+Gazebo
plugin to provide telemetry from other important UAS payload sensors
(\S~\ref{subsec:Core-Components}).
\end{enumerate}
These three components work together in real time to enable a UAS
simulation of hypothetical (or real) UAS systems that have sensor
imaging capabilities outside (and inside) the visible frequency spectrum,
e.g., IR, radio-frequency (RF).

\subsection{\label{subsec:Google-Earth-Engine}Google Earth Engine client}

The Google Earth Engine client performs a number of tasks that would
be time-consuming and prone to error if performed manually. Specifically,
it automates the time-consuming process of aggregating and downloading
the requested image data, merging image tiles, outputting the required
text files and populating these files with the correct information
to create a metrically accurate representation of the scene as a Gazebo
model. The google earth engine client requires users to specify the
following inputs:
\begin{itemize}
\item The image data set of interest,
\item The location, (latitude,longitude), of the image center and the dimension
of the desired image (in meters),
\item Any data set search refinement parameters, e.g., cloudy/clear visibility
days, time of year, etc.
\end{itemize}
From these parameters, the Earth Engine client will access Google's
image database system and create the required service requests to
obtain the image data and generate the Gazebo model.

Google's Earth Engine system allows researchers to use software tools
to search massive data repositories \cite{citeEarthEngine} to make
new discoveries. While our discussion is limited to describing multi-spectral
applications, we see a wide possibility of applications for the software
using alternative data set types. Another desirable aspect of these
data are the annotations that add value for selecting specific measurement
contexts which allow researchers to find images based on the date/time-of-day
(Sentinel 1-3), see table \ref{tab:Multi-spectral-image-data-1},
atmospheric conditions, e.g., ``cloudy'' \cite{citeG,citeM}\textbf{
}and many other useful criteria such as quality, time of day, grazing
angle, etc., see table \ref{tab:Derived-image-data-1}. Note, each
entry from the following tables can be accessed via the Earth Engine
website \cite{citeEarthEngine} by searching for the associated `Name'
entry.\textbf{ }
\begin{table}[H]
\begin{centering}
\begin{tabular}{|c|c|c|}
\hline 
Name & Wavelengths & Resolution\tabularnewline
\hline 
\hline 
skysat  & RGB & 0.8m\tabularnewline
\hline 
NAIP  & RGB & 1.0m\tabularnewline
\hline 
landsat-8  & 11 wavelengths & 30m\tabularnewline
\hline 
Sentinel-1  & 5.4GHz 4 polarities & 10m\tabularnewline
\hline 
Sentinel-2  & 11 wavelengths & 20m\tabularnewline
\hline 
Sentinel-3  & 21 wavelengths & 0.0139465 $W/m^{2}sr\mu m$\tabularnewline
\hline 
\end{tabular}
\par\end{centering}
\caption{\label{tab:Multi-spectral-image-data-1}Multi-spectral image data,
\cite{citeEarthEngine}}
\end{table}

Table \ref{tab:Multi-spectral-image-data-1} provides a subset
of the multi-spectral image data sources available from Google Earth
Engine. Creation of multiple ROSgeoregistration nodes for the same
3D planar model allows researchers to simulate sensing UAS sensor
payloads that currently reside only on sophisticated satellite systems.
Simulation tools allow researchers to explore the potential classification,
navigational and operational capabilities these sensors can provide
for more advanced UAS systems.\textbf{}
\begin{table}[H]
\begin{centering}
\begin{tabular}{|c|c|c|}
\hline 
Name & Units & Range\tabularnewline
\hline 
\hline 
Hourly Surface temp. & Kelvin & 223.6$^{\circ}$- 304$^{\circ}$\tabularnewline
\hline 
Monthly Precipitation  & m & 0 - 0.02m\tabularnewline
\hline 
Monthly Wind  & m/sec & $\pm$11.4$^{\circ}$\tabularnewline
\hline 
Vegetation  & classes & 584 types\tabularnewline
\hline 
Elevation  & m & .33 arc seconds\tabularnewline
\hline 
\end{tabular}
\par\end{centering}
\caption{\label{tab:Derived-image-data-1}Derived image data, \cite{citeEarthEngine}}
\end{table}

Table \ref{tab:Derived-image-data-1} provides a subset of additional
image data sources that can also be georegistered and used in conjunction
with the aforementioned spectral data. These metadata products could
provide key attributes for more reliable recognition or key insights
on measurement data to discover unknown correlations between scene
reflectance and other phenomena.\textbf{}

\subsection{\label{subsec:ROS+gazbeo-plugin}ROS+gazbeo plugin}

To maximize the benefit of prior work, ROSgeoregistration, depends
on the core ROS UAS simulation tools. Specifically, ROSgeoregistration
uses the core as a mechanism to create a vehicle with an arbitrary
sensor payload and navigate that vehicle in a simulated 3D environment.
ROSgeoregistration then provides additional simulation telemetry that
is highly-valuable for conceiving, developing and evaluating algorithms
that may seek to leverage recorded sensor image data from the UAS
platform for pose or derivative, e.g., velocity or acceleration estimation.

\subsubsection{ROSgeoregistration parameters}

ROSgeoregistration is a plugin for Gazebo that allows multiple textures
to be mapped to a polygonal model within the Gazebo world. Each instance
of the plugin creates an application thread, also referred to as a
ROS node, that uses the current sensor pose and image formation parameters
to generate image telemetry.

Each downloaded terrain model is linked into the Gazebo world (.world
file) using an ``include'' directive to place the model geometry
and texture into the 3D Gazebo scene. An instance of the simulation
plugin is required to generate a simulated view of the terrain model.
Hence, simulation of simultaneous EO, IR, and RF imagery of the terrain
requires (3) threads where each thread uses the same model geometry
but distinct reference image textures.

Telemetry is generated by finding the projective transformation that
maps pixels in the reference image to the pixels of the sensor image.
This is accomplished via geometric optics. Specifically, we use the
relative geometry of the terrain model and the sensor pose in conjunction
with the sensor/camera image formation parameters (aperture, focus,
distortion, etc) to determine the world $(X,Y,Z)$ coordinates measured/viewed
by the sensor. 

\subsubsection{Simulating Sensed Images}

As with most image sensors, our simulator uses a rectangular image
sensor model for image formation whose measurements are recorded on
a discrete rectangular grid where each pixel denotes a rectangular
cell of this grid. The rectangular sensor geometry imposes a pyramidal
view frustum into the simulated 3D environment and our simulator computes
the intersection of this view frustum with the 3D planar model containing
the reference image data. The intersection locations are used to establish
a correspondence, in the form of a projective transformation, between
the reference image and the sensed image. 

In general, the eight unknown values of the perspective transformation
between the reference image and the sensed image require four correspondences
between image points and their corresponding $(X,Y,Z)$ locations
from the terrain model. The four correspondences are sufficient to
solve for the unknowns of the projective transformation by solving
a linear system (or by using least-squares when more than four correspondences
are provided). This task is common in computer vision image matching
applications such as stereo 3D reconstruction. Hence we use the solution
to this problem provided by the OpenCV library made available via
the \emph{cv::getPerspectiveTransform()} function.

\subsubsection{Sensor-Terrain Model Boundary Conditions}

While any four image points will suffice to estimate the perspective
transformation, we use the corners of the image to detect when the
view of the sensor does not completely fall within the reference terrain
model. This can occur when the vehicle flight path causes the sensor
view to deviate from the reference image model and typically occurs
when the vehicle is not flying over the reference image plane or when
vehicle path-following maneuvers, e.g., rolls or pitches, cause the
sensor view to exceed the boundary of the terrain reference model.
In cases when the entire sensor image cannot be simulated, the simulator
does not output image telemetry.

\subsubsection{Coordinate Systems}

All coordinates from ROSgeoregistration originate from Google Earth
Engine. As such, the local $(X,Y,Z)$ world coordinate system is directly
linked to standard geospatial coordinate systems, e.g., EPSG:4326
(the Mercator projection) and EPSG:3857 (the Web Mercator projection).
With proper parameter specification this allows local $(X,Y,Z)$ values
to transfer seamlessly to geospatial coordinate frames, e.g., (latitude,
longitude). This is typically easily accomplished by collocating the
origin with the center of the georeference source image data.

\subsection{\label{subsec:Core-Components}Core Components}

To deploy ROSgeoregistration, one must first install and configure
the simulator ``core'' dependencies. The ROSgeoregistration packages
use the core components to provide real-time simulated sensor image
data that is consistent with other simulated vehicle telemetry. This
allows the simulated ground truth projective transformations and ground
truth sensor poses generated to be fused with other sensor telemetry
from the simulated UAS flight. The simulator core components consist
of the following ROS packages:
\begin{itemize}
\item ROSflight+ROSflight plugins \cite{Rosflight}: a suite of ROS packages
to allow UAS flight. Plugins add simulated UAS sensor telemetry.
\item ROSplane \cite{Rosplane}: a suite of ROS packages to allow flight
of real or simulated fixed wing aircraft.
\item ROScopter \cite{ROScopter}: a suite of ROS packages to allow flight
of real or simulated multi-rotor drones, e.g., quadcopters.
\end{itemize}
Collectively the core components allow users to simulate a UAS realized
as either a fixed wing or multi-rotor aircraft. The aircraft will
be placed in a Gazebo virtual environment which uses the bullet physics
engine to simulate standard rigid-body Newtonian forces, e.g., friction,
collisions, gravity, inertia, etc. The ROSflight package and, depending
on the vehicle, ROSplane or ROScopter packages provide additional
forces including lift, wind forces and flight dynamics associated
with the chosen vehicle model.

\subsubsection{Flight dynamics simulation with ROSflight}

The ROSflight software described in \cite{Rosflight} provides a suite
of ROS packages. For simulation, ROSflight adds forces of flight dynamics
into the simulator for both fixed wing aircraft and for muti-rotor
drones, e.g., quadcopters. The ROSflight also has plugins that simulate
telemetry from important sensors including GPS, airspeed, barometric
pressure, an Inertial Measurement Unit (IMU) and a magnetometer. Also
included are XML format files (.xacro files) that allow client applications
to easily incorporate these sensors into their virtual robot representation
by including these sensors in their specific Universal Robot Description
File (URDF).

\subsubsection{Fixed wing aircraft simulation with ROSplane}

The ROSplane ROS package described in \cite{Rosplane} provides path
planning, path following algorithms and the associated state estimators,
e.g., Extended Kalman Filters (EKFs), and controllers required to
fly simulated or real fixed wing aircraft. In addition, ROSplane provides
enhanced simulation components that incorporate additional forces
into the simulation environment and also publish the ground truth
state of the fixed wing aircraft.

\subsubsection{Quadcopter or drone simulation with ROScopter}

The ROScopter ROS package described in \cite{ROScopter} provides
path planning, path following algorithms and the associated state
estimators and controllers required to fly simulated or real multi-rotor
aircraft. In addition, ROScopter provides enhanced simulation components
that incorporate additional forces into the simulation environment
and also publish the ground truth state of the multi-rotor aircraft.

\section{\label{sec:Results}Results and Future Work}

For results, we describe the new outputs made available by ROSgeoregistration
and discuss a use-case experiment where we apply this tool to view
a terrain model represented in both the EO (visible light; 400\textendash 790
THz) and RF (C-band; 5.4GHz) frequency ranges. 

\subsection{Experimental Setup}

In our example experiment the user wishes to develop, design and evaluate
different algorithms for image-based guidance, control and dynamics
(GNC) using real-world EO and SAR image data sourced from Google Earth
Engine. Generically, a user must complete the steps listed below to
run a simulation:
\begin{enumerate}
\item Choose a latitude and longitude location of interest and size for
the terrain model,
\item Choose the specific Google Earth Engine (or Google Maps) source data
and data set parameters for analysis,
\item Use the \textbf{Google Earth Engine client} to locate, download and
construct a Gazebo model of the scene as a collection of georeferenced
images,
\item For each sensor of each vehicle define the sensor mounting locations
and the sensor image formation model,
\item For each vehicle, choose a vehicle a flight path and navigational
controller to traverse the path. 
\item \textbf{Run the ROSgeoregistration simulator.}
\item Image-based GNC algorithms take in images and compute GNC relevant
outputs, e.g., vehicle sensor pose.
\end{enumerate}
In the following paragraphs we detail our path through these steps.

In steps (1,2) we use the UNC Charlotte ECE building roof (35.395703N,
-80.535865E) as the UAS launch site and downloaded Google Earth EO
satellite data (as provided in Google Maps) and Sentinel-1 SAR data.

In step (3) the Google Earth Engine client merged EO image from 100
high-quality and unobstructed (not cloudy) EO image tiles to generate
a $(x,y)$ image with resolution of 6400$\times$6150 pixels spanning
a 7643m.$\times$7345m. region ($\sim$1m/pixel resolution). The client
extracted Sentinel-1 SAR data (see table \ref{tab:Multi-spectral-image-data-1})
for the same region of the Earth's surface generating a $(x,y)$ image
with resolution 765$\times$600 pixels spanning the same region ($\sim$10m/pixel
resolution).

In steps (4,5) we selected a fixed-wing aircraft platform and mounted
a virtual downward looking RGB sensor and SAR sensor by modifying
the robot URDF. We then used the standard path following controller
and path management software from ROSplane to control the flight surfaces
and propulsion to traverse a sequence of way-points.

\begin{figure}
\begin{centering}
\subfloat[]{\begin{centering}
\includegraphics[height=2.5cm]{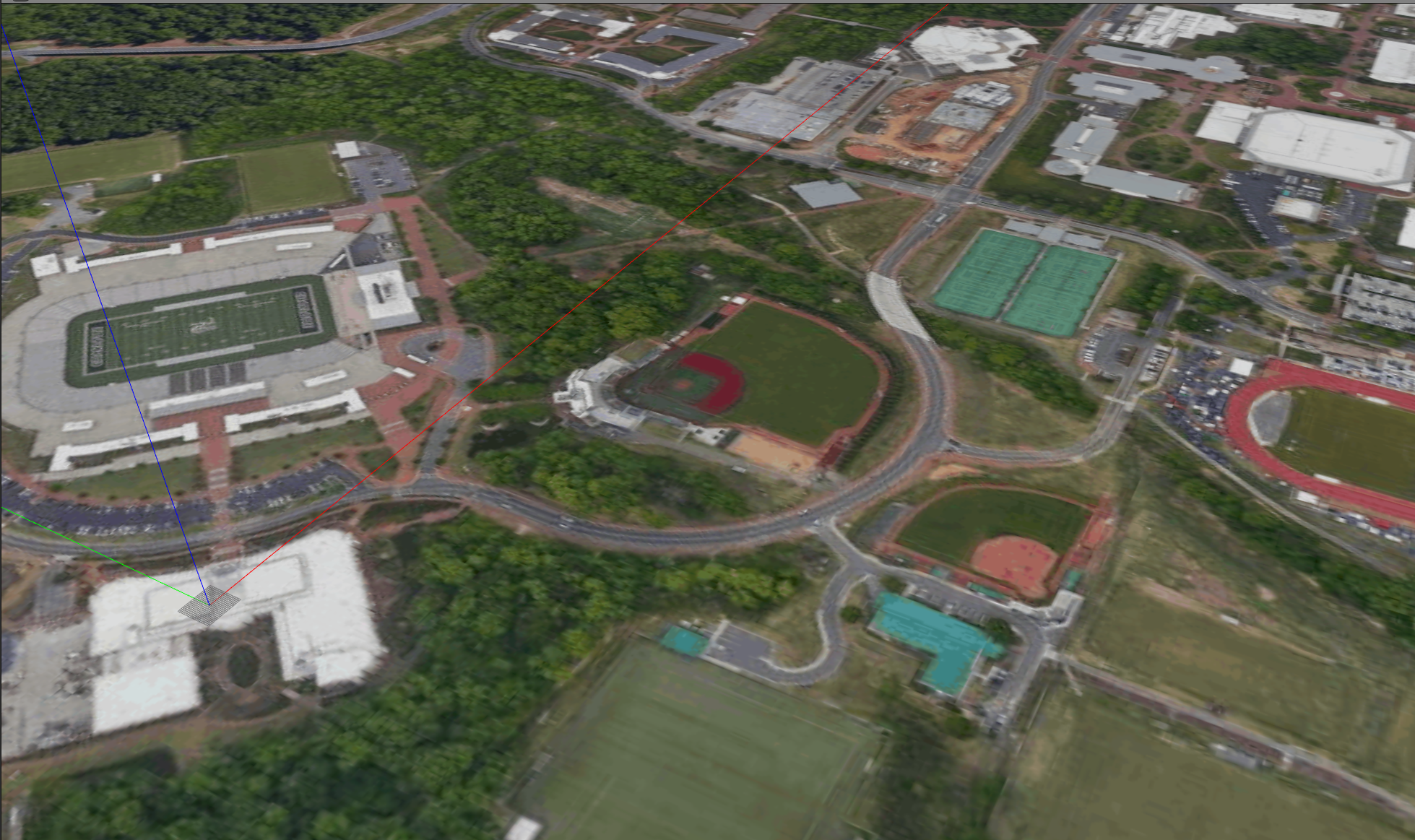}
\par\end{centering}
}\subfloat[]{\begin{centering}
\includegraphics[height=2.5cm]{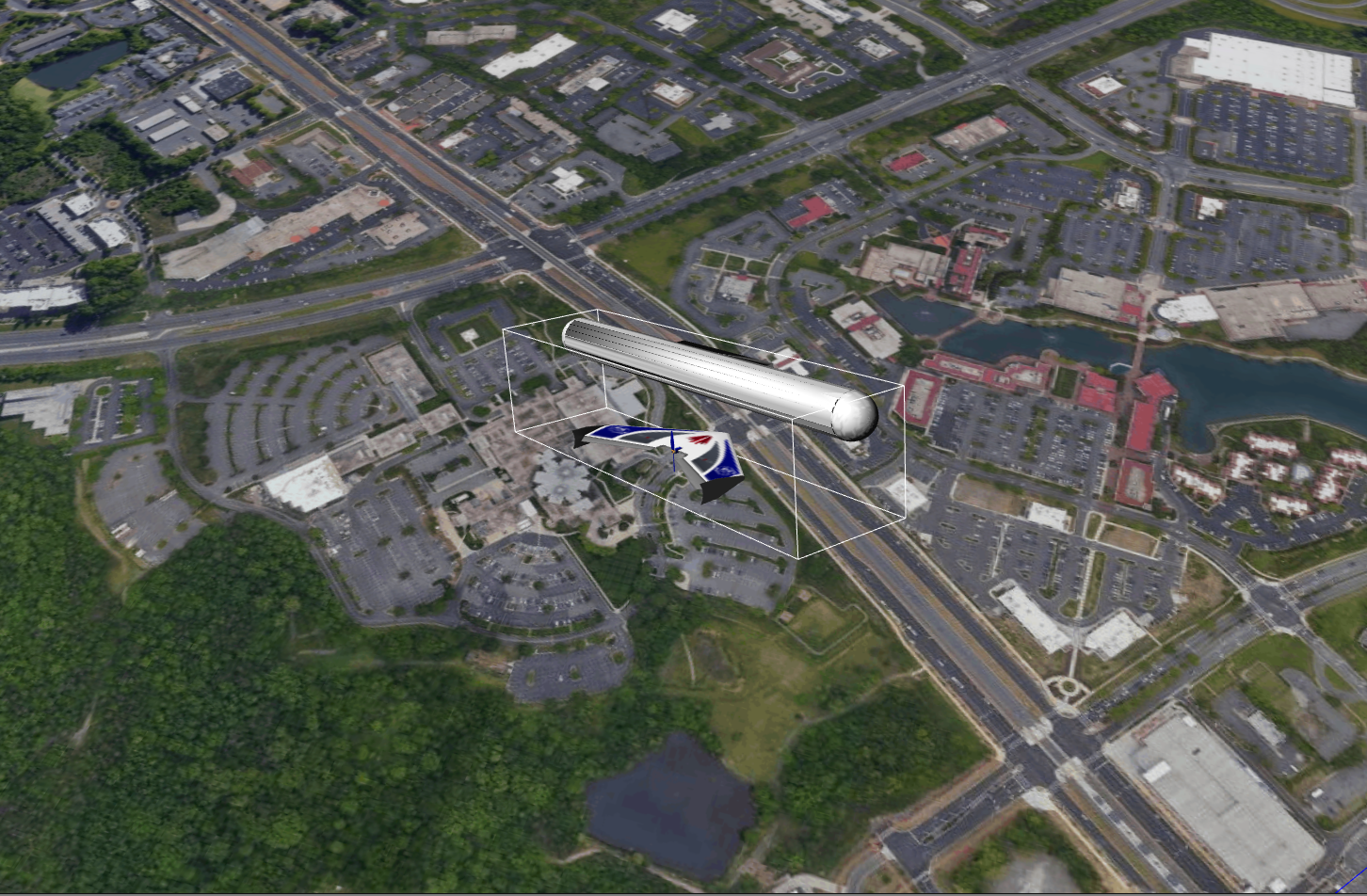}
\par\end{centering}
}
\par\end{centering}
\caption{\label{fig:EO}(a,b) show views of the simulation scene as rendered
by the Gazebo simulator. }
\end{figure}

\begin{figure}
\begin{centering}
\subfloat[]{\begin{centering}
\includegraphics[height=4cm]{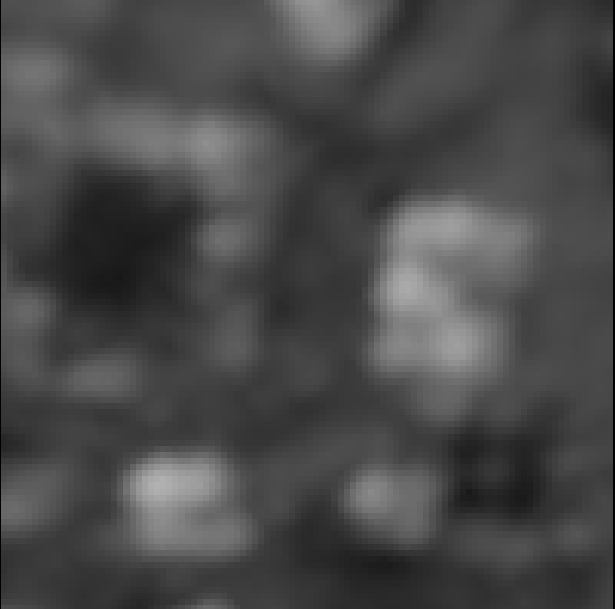}
\par\end{centering}
}\subfloat[]{\begin{centering}
\includegraphics[height=4cm]{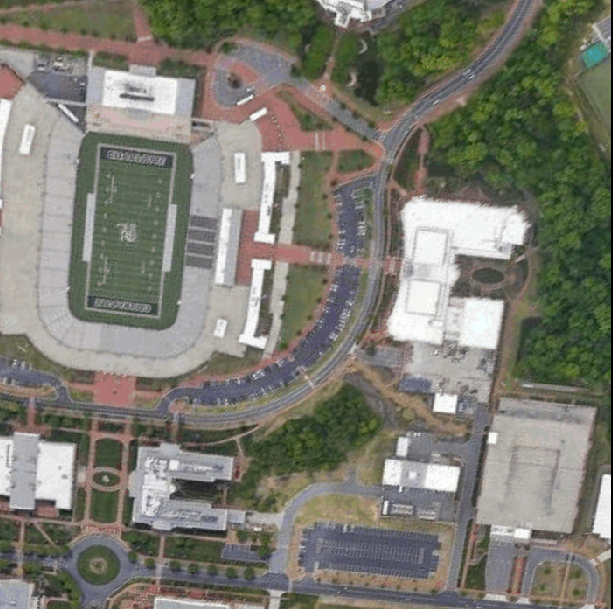}
\par\end{centering}
}
\par\end{centering}
\caption{\label{fig:Simulated-SAR-and-EO}(a,b) shows simulated images of the
same spatial region of the terrain model as it would appear for a
SAR sensor and an EO sensor. The structure in the top left of the
image is the UNC Charlotte 49ers football stadium. Note image (a)
has 10m/pixel resolution and (b) has 1m/pixel resolution. }
\end{figure}

In step (6) we modified the Gazebo world to included the terrain models
and added the required ``node'' elements to the xml of the ROS simulation
``launch'' file. This file initiates the necessary commands to start
the simulation and the ROSgeoregistration plugins to simulate the
EO and SAR image data as the vehicle flies through the simulated environment.
A rendering of the simulation scene, and one showing the UAV in the
Gazebo environment are shown in figure \ref{fig:EO}. A typical sensor
rendering for both EO and SAR are shown in \ref{fig:Simulated-SAR-and-EO}.

\begin{figure}
\begin{centering}
\subfloat[]{\begin{centering}
\includegraphics[height=3.5cm]{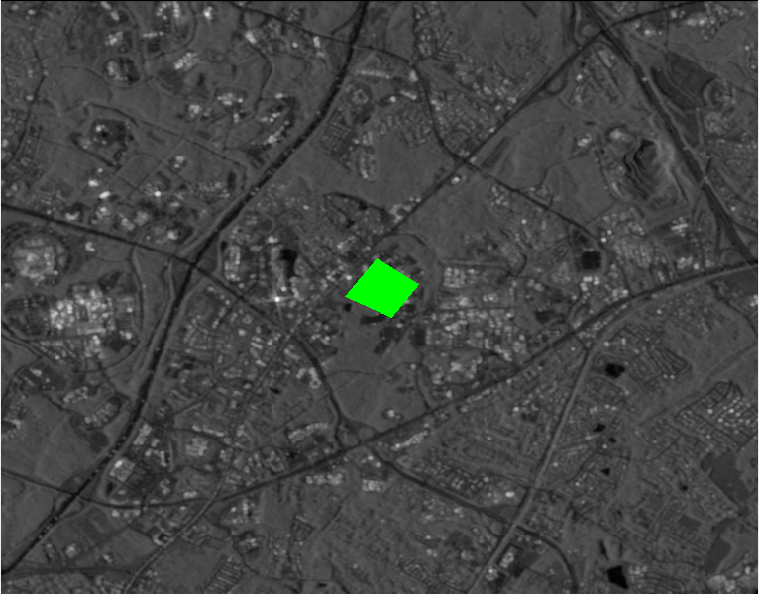}
\par\end{centering}
}\subfloat[]{\begin{centering}
\includegraphics[height=3.5cm]{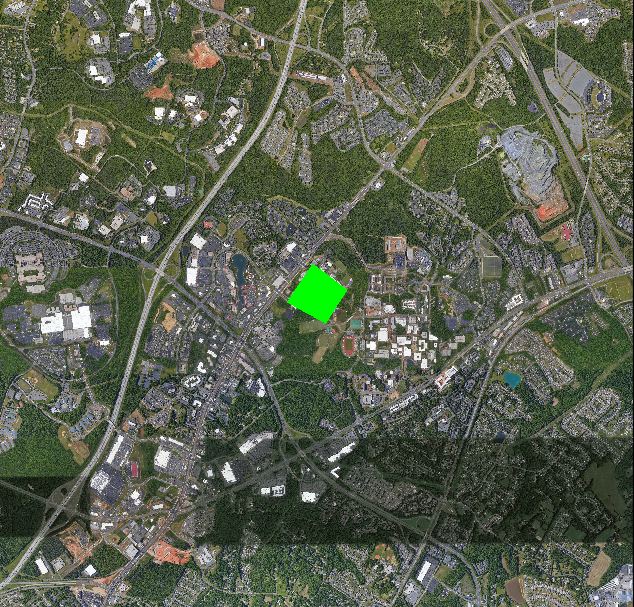}
\par\end{centering}
}
\par\end{centering}
\caption{\label{fig:footprint}(a,b) show pixel regions taken from the SAR
and EO reference images respectively to form the images of fig. \ref{fig:Simulated-SAR-and-EO}.
The sourced pixel regions are shown as green quadrangles.}
\end{figure}

In step (7) we apply the simulated images to evaluate the performance
of two image registration algorithms. Fig. \ref{fig:Simulated-SAR-and-EO}
shows corresponding views of the same spatial region in the RF band
(a) and EO band (b). The stadium and building structures are seen
to be in correspondence. Fig. \ref{fig:footprint}(a,b) show image
regions extracted from SAR and EO terrain model to generate the images
of Fig. \ref{fig:Simulated-SAR-and-EO}.
 
\subsection{Example Application}

We apply the simulator to evaluate the performance of two registration
algorithms: (1) feature-based registration and (2) mutual-information
registration. Our application uses these algorithms to match sensed
EO and SAR images to a georeferenced EO image stored on the UAS. Registration
results can then be used in conjunction with camera calibration parameters
to estimate the sensor/vehicle pose (see \cite{Hartley:2003:MVG:861369}).

\begin{figure}
\begin{centering}
\subfloat[]{\begin{centering}
\includegraphics[height=2.7cm]{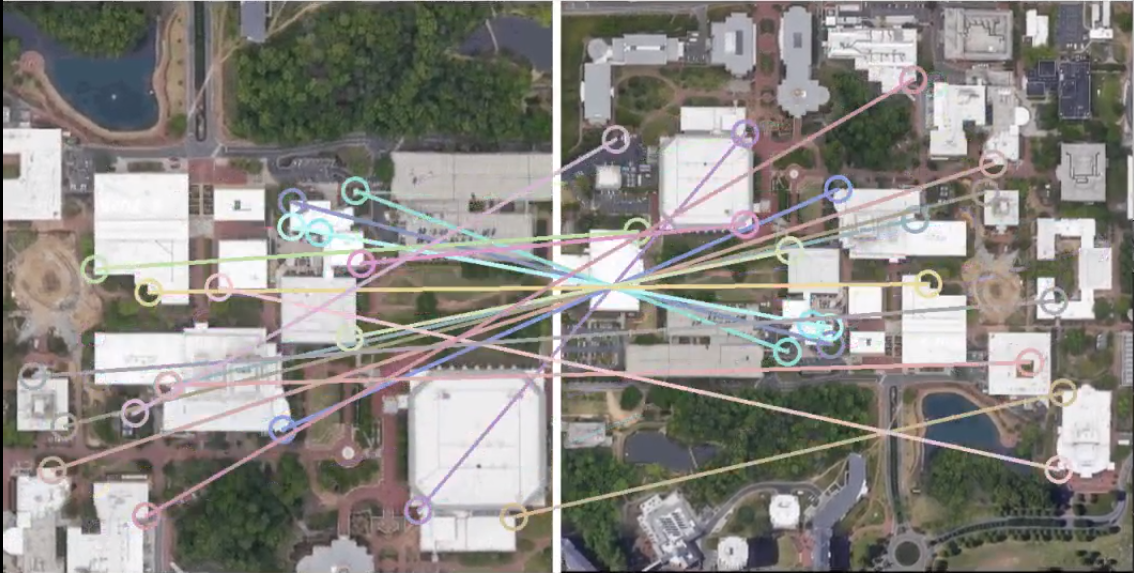}
\par\end{centering}
}\subfloat[]{\begin{centering}
\includegraphics[height=2.7cm]{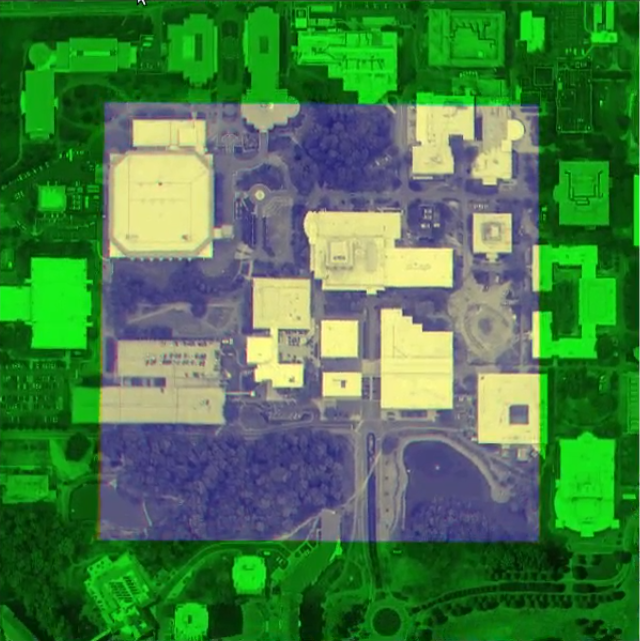}
\par\end{centering}
}
\par\end{centering}
\begin{centering}
\subfloat[]{\begin{centering}
\includegraphics[height=2.7cm]{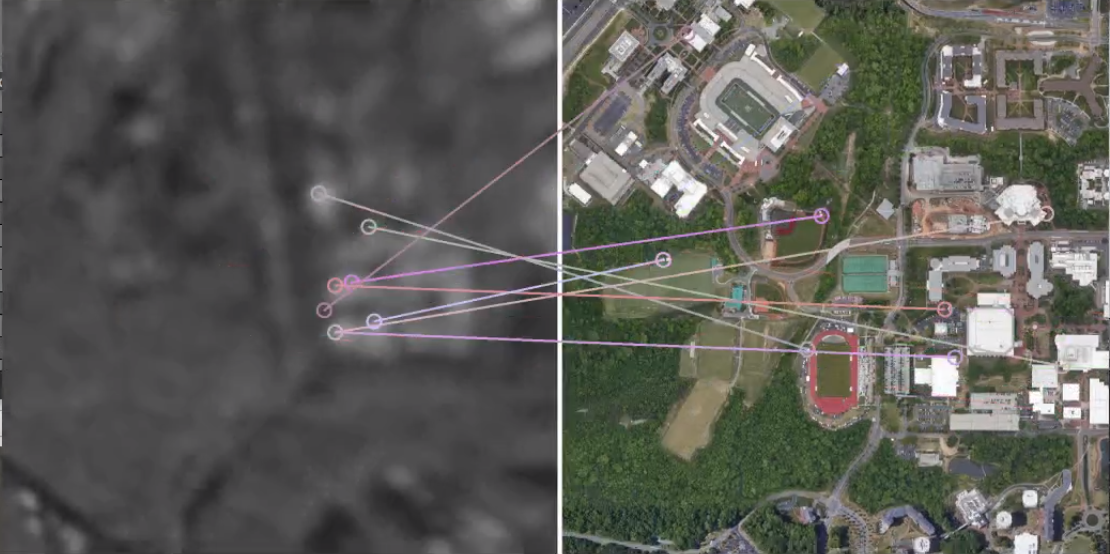}
\par\end{centering}
}\subfloat[]{\begin{centering}
\includegraphics[height=2.7cm]{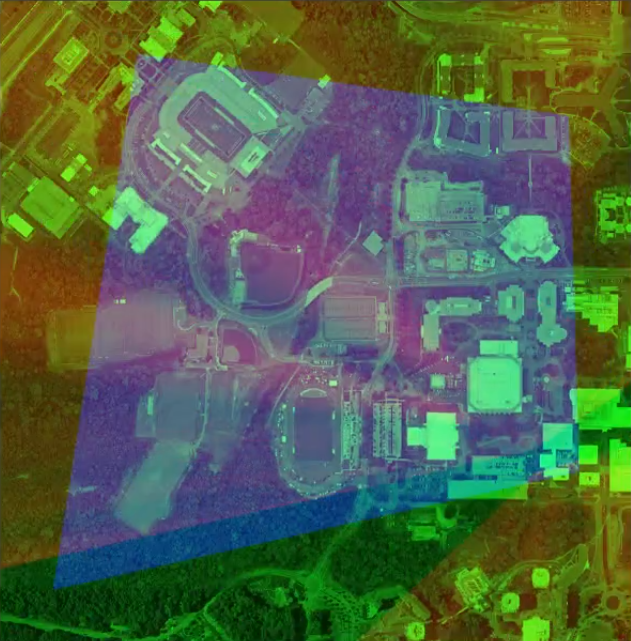}
\par\end{centering}
}
\par\end{centering}
\begin{centering}
\subfloat[]{\begin{centering}
\includegraphics[height=2.7cm]{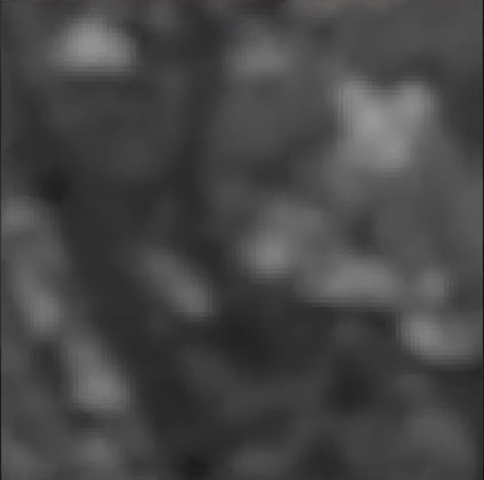}~\includegraphics[height=2.7cm]{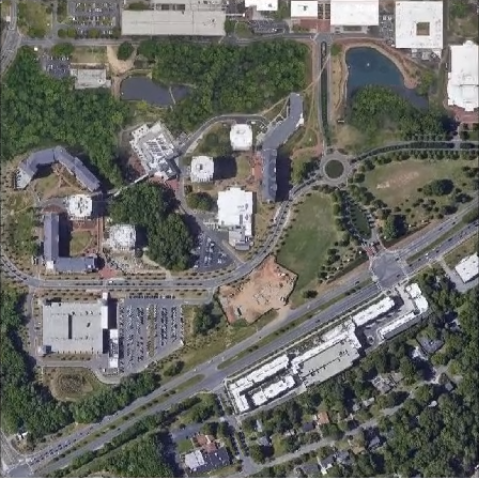}
\par\end{centering}
}\subfloat[]{\begin{centering}
\includegraphics[height=2.7cm]{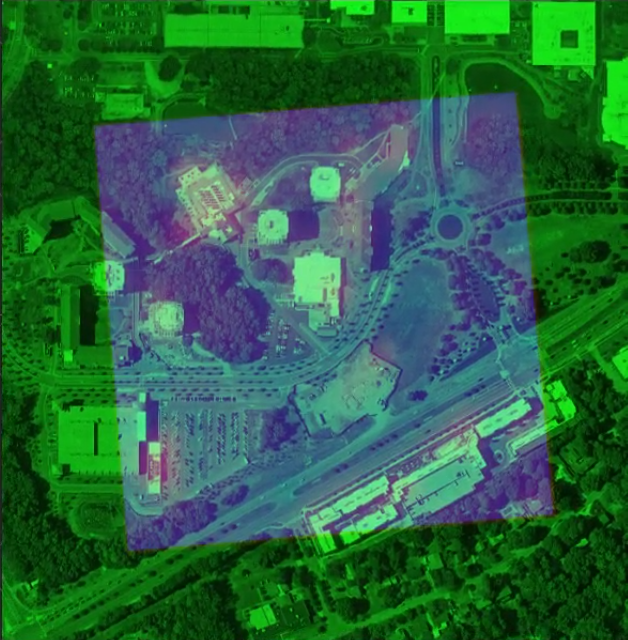}
\par\end{centering}
}
\par\end{centering}
\caption{\label{fig:Homography-Results}(a-f) show the application of our simulator
to assessing SAR-and-EO image registration algorithms. (a,c,e) shows
a pairs of images matched in a registration application. (b,d,f) show
visual assessments of the accuracy of these matches using the ground
truth perspective transforms available from the simulator. }
\end{figure}

Figure \ref{fig:Homography-Results} shows homography results where
camera/sensor image (left) is matched to a portion of the onboard
georeferenced EO image. Effectively the inverse problem of generating
the image in the first place, but without access to truth data. In
this case, ORB features \cite{Rublee2011} were matched using RANSAC
to find EO-to-EO image correspondences and solve the projective transform
that maps pixels of the camera image to the locations in georeferenced
EO image \cite{Goshtasby2005}. The quality of the solution is depicted
as a fused RGB image where the image components are (red=matched intensities
sensed EO image, green=georeferenced EO intensities, blue=quadrangular
region of the ground truth homography). We find that feature-based
matching works well for EO-to-EO imagery having similar scales (see
Fig. \ref{fig:Homography-Results}(a,b)). However, this approach fails
when used on SAR-to-EO matching problems (see Fig. \ref{fig:Homography-Results}(c,d)).
In contrast, the mutual information registration approach is more
successful in estimating SAR-to-EO homographies (see Fig. \ref{fig:Homography-Results}(e,f)).

In addition to the simulated image data, the ROSgeoregistration publishes
important additional data to client applications to facilitate ground
truth based assessment by client applications. This information includes
the camera intrinsic parameters, the camera pose, $(X,Y,Z)$ scene
footprint (corner) locations of the camera aperture/view, and $(u,v)$
locations of the footprint within the planar geometry, i.e., analogous
to $(u,v)$ texture coordinates. 

\subsection{Future Work}

One clear drawback of the current approach is the lack of terrain
elevation. While this is not a significant concern for many high altitude
UAV applications including those likely to leverage SAR as in our
example, it does limit the realism for lower altitude flights. Tying
vision reference maps into digital terrain elevation data (DTED) would
require additional computation when generating camera images, but
would result in more realistic view irregularities for camera image
to reference map registration. 

While the tool currently produces SAR imagery commensurate with a
well known trajectory, additional functionality could be added to
the SAR image retroactively to approximate effects of navigation error
\cite{SARerror}, or moving targets \cite{Sar_addTarget}, as mentioned
previously. This manipulation could be done independently to support
SAR-to-EO matching, etc., or in-the-loop with image modification being
a function of the partnered navigators current state errors. \textcolor{red}{}

\section{\label{sec:Conclusion}Conclusion}

This article introduces a new simulation tool, ROSgeoregistration,
to simulate high-altitude, multi-spectral aerial image sensor data.
The software represents a novel integration of Google's Earth Engine
with the ROS+Gazebo 3D simulation environment to provide new tools
to UAS researchers. Our client to Google Earth Engine greatly reduces
the complexity of leveraging the massive multi-spectral imaging database
of Google's Earth Engine to generate terrain models usable by the
Gazebo simulator by automatically downloaded the needed data, addressing
the coordinate frame transformations, and stitching together image
tiles as needed for large reference maps/simulation environments.
Our ROS+Gazebo plugin simulates geospatially and radiometrically accurate
imagery from multiple sensor views of the same terrain region, and
does so accurately, i.e., EO and SAR reference images (and their simulated
camera/sensor outputs) are self-consistent. Finally, EO and SAR image
generation capabilities were developed to integrate with other ROS
vehicle simulation tools provide rich telemetry data, e.g., IMU, GPS,
air speed and magnetometer, for data fusion applications. It is hoped
that the combination of these tools in one centralized toolkit provides
researchers with relevant data for a range of R\&D activities including
image and multi-spectral processing, navigation, and much more. 

\section{\label{sec:Acknowledgments}Acknowledgment}

This research is sponsored by an AFRL/National Research Council fellowship
and results are made possible by resources made available from AFRL's
Autonomous Vehicles Laboratory at the University of Florida Research
Engineering Education Facility (REEF) in Shalimar, FL.

\bibliographystyle{ieeetr}
\bibliography{2022_ICUAS_Aerial_SAR_Simulator}

\end{document}